%% file: main.tex
\pdfoutput=1
\documentclass[10pt, logo, twocolumn, copyright]{nvidiatechreport}
\usepackage{caption}
\usepackage[utf8]{inputenc}
\usepackage[T1]{fontenc}
\usepackage{hyperref}
\usepackage{url}

\usepackage{amsfonts}
\usepackage{microtype}
\usepackage[dvipsnames]{xcolor}
\usepackage{graphicx}
\usepackage{subcaption}
\usepackage{placeins}
\usepackage{booktabs}
\usepackage{tabularx}
\usepackage{multirow}
\usepackage[numbers,sort&compress]{natbib}
\usepackage{xspace}
\usepackage{amsmath}
\usepackage{listings}
\usepackage{enumitem}


\title{Heterogeneous Parallelism for Multimodal Large Language Model Training}
\author{
Yashaswi Karnati,
Kamran Jafari,
Akash Mehra,
Li Ding,
Pranav Prashant Thombre,
Ali Roshan Ghias,
Shifang Xu,
Parth Mannan,
Yu Yao,
Hao Wu,
Eric Harper,
Ashwath Aithal,
Nima Tajbakhsh
}
\correspondingauthor{}

\begin{abstract}
\textbf{Abstract:}
Foundation model training is becoming multimodal, from post-training pipelines to large-scale pretraining. As modality coverage broadens, context windows grow, and encoder–LLM scales diverge, a single LLM-centric TP/CP/PP/DP/EP layout increasingly limits throughput. This coupling forces encoders to inherit LLM-driven sharding and placement choices that can add communication, limit encoder parallelism, or constrain the LLM schedule; the mismatch is most pronounced at long contexts, where LLM context parallelism is needed for the fused multimodal sequence but encoder inputs remain bounded. We present heterogeneous parallelism for multimodal large language model training, an abstraction that lets modules in one end-to-end graph use independent layouts and rank placements, supporting colocated execution on shared GPUs and non-colocated execution on disjoint rank sets. The key challenge is preserving boundary tensor semantics across independent layouts: forward activations must be materialized for the destination layout, while backward gradients must be routed back to the source layout. We address this with boundary communicators that implement forward and backward layout transforms, plus scheduling extensions for both placement modes. We evaluate optimized homogeneous, colocated heterogeneous, and non-colocated heterogeneous configurations across multimodal workloads and GPU scales to characterize when added layout and placement freedom exposes a better operating point. Across this sweep, colocated heterogeneity improves TFLOPS/GPU by up to 49.3\%, while non-colocated heterogeneity improves aggregate token throughput by up to 13.0\% and TFLOPS/GPU by up to 9.6\%. We validate loss convergence parity against homogeneous baselines and release the system as an open-source Megatron-LM extension.
\end{abstract}

\begin{document}
{\hbadness=10000\maketitle}

\input{tex/introduction.tex}
\input{tex/related.tex}
\input{tex/method.tex}
\input{tex/experiments.tex}
\begin{figure}[!t]
  \centering
  \includegraphics[width=\linewidth]{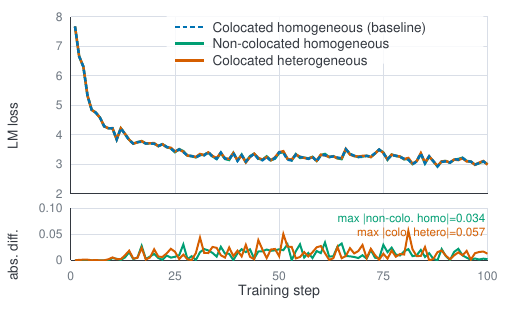}
  \caption{Controlled FP32 convergence comparison. Baseline:
  colocated homogeneous on four GPUs with the vision encoder and LLM both
  using \(\mathrm{TP}=4,\mathrm{DP}=1\); non-colocated same-layout keeps
  \(\mathrm{TP}=4,\mathrm{DP}=1\) for both modules but places the LLM on ranks
  0--3 and the vision encoder on ranks 4--7. Colocated heterogeneous keeps the
  same four ranks but uses vision \(\mathrm{TP}=1,\mathrm{DP}=4\) and LLM
  \(\mathrm{TP}=4,\mathrm{DP}=1\); both variants follow the baseline loss
  curve.}
  \label{fig:convergence-homo-hetero}
\end{figure}
\input{tex/convergence_validation.tex}

\section{Conclusion}
\label{sec:conclusion}

We presented heterogeneous parallelism for multimodal training, a runtime
abstraction that lets modules in one end-to-end graph use independent layouts
and rank placements while boundary communicators and scheduling extensions
preserve activation and gradient semantics across colocated and non-colocated
pipeline execution. This flexibility lets the training system tune its operating
point as model scale, context length, modality-token density, memory pressure,
communication volume, and pipeline balance change. In our evaluated workloads,
colocated execution is effective
when the same GPU allocation can support different module grids, whereas
non-colocated execution becomes useful when encoder state on the LLM ranks
restricts the LLM memory or pipeline layout. Across these workloads, this
yields up to 49.3\% TFLOPS/GPU for colocated execution and 13.0\% aggregate
token throughput for large-encoder 120B LLM training.
FP32/BF16 validation confirms training behavior. The open-source Megatron-LM
extension already supports multiple encoders, audio/video, and MoE; future work
will evaluate their performance regimes, including how extra encoders and module
boundaries shift layout and placement choices.
\clearpage

\begingroup
\small
\raggedright
\bibliographystyle{abbrvnat}
\bibliography{paper}
\endgroup

\input{tex/appendix.tex}

\end{document}

%% file: tex/introduction.tex
\section{Introduction}
\label{sec:intro}

As foundation-model training expands from text-only transformers to multimodal systems spanning images, audio, video, and documents, the training graph has grown heterogeneous while the parallelism stack has not kept pace. The resulting graph consists of modules with fundamentally different depth, width, and context size — and therefore different requirements for TP, CP, PP, DP, and EP\footnote{TP, CP, PP, DP, and EP denote tensor, context, pipeline, data, and expert parallelism, respectively.} — yet in practice all modules are constrained to a single shared layout. While choosing this layout for the LLM\footnote{LLM denotes large language model; MLLM denotes multimodal large language model.} is natural, encoder and LLM parallelism requirements increasingly diverge at scale.

These mismatches motivate heterogeneous multimodal parallelism: layout should be chosen per module rather than as a model-wide constant.
\emph{First}, the high TP degrees required by large LLMs introduce
all-reduce collectives around otherwise local encoder computation, and this
overhead can outweigh the compute and memory savings for small encoders.
\emph{Second}, and most consequentially, CP shards the fused language
sequence to manage attention memory, but encoder sequence length is bounded by
modality resolution rather than the LLM context window. As a result, inheriting
LLM CP shards a fixed-size encoder sequence without memory relief, while those
ranks could instead act as independent encoder DP replicas and process more
modality samples per step. \emph{Third}, PP confines encoder execution to the
first PP stage of LLM, forcing the encoder to inherit that stage's TP degree and encoder-heavy workloads can overload those ranks and prevent the LLM from using a more efficient pipeline layout.

Which mismatch matters most is workload-dependent. Changing encoder and LLM
scale, context length, modality-token density, world size, or optimizer-state
sharding shifts the balance among compute, activation memory, optimizer-state
footprint, communication volume, and pipeline balance. Heterogeneous
parallelism therefore does not prescribe one universally better layout; it
exposes additional degrees of freedom so the system can tune encoder and LLM
layouts to the workload's operating point.

Decoupling module layouts, however, is not just a configuration change: adjacent modules must still agree on boundary tensor semantics. Forward activations must be materialized in the destination layout, and backward gradients must be returned to the source layout. This requirement raises three system challenges. \emph{First}, every inter-module edge needs a paired forward and backward layout transform. The central correctness challenge is that an incorrect layout mapping can produce tensors with valid shapes, allowing training to proceed while silently routing gradients to the wrong ranks and corrupting optimization without a runtime error. \emph{Second}, colocated
placement requires each rank to interpret its position in two logical grids and
apply the correct forward layout transform, with the matching reverse transform
in backward. \emph{Third}, pipeline scheduling must generalize beyond a single
linear stage chain: multimodal graphs can contain modules with different PP
depths and boundary edges that require communicators distinct from ordinary
pipeline P2P. Solving all three together, without changing the model or training semantics, is the core technical problem this paper addresses.

We present heterogeneous parallelism for multimodal large language model
training as an open-source Megatron-LM extension that lets each module choose
its own TP/CP/PP/DP/EP layout and physical rank set while preserving a single
end-to-end training graph. Its core primitive is the boundary communicator, an
autograd-aware operator attached to an inter-module edge: in forward, it
materializes the source activation in the destination module's layout; in
backward, it applies the inverse transform so the gradient returns to the source
module's layout. The same abstraction supports two complementary placement
modes. In colocated execution, modules share physical GPUs but interpret those
ranks under different logical grids, which is useful when encoder and LLM can
share hardware but should not share a layout. In non-colocated execution,
modules occupy disjoint rank sets, which is useful when encoder state on the LLM
ranks restricts the LLM memory or pipeline layout.

Scheduling extensions compose these boundary transforms with pipeline-parallel
training. A graph-aware 1F1B\footnote{1F1B denotes one-forward-one-backward
pipeline scheduling; P2P denotes point-to-point communication.} dispatcher
coordinates non-colocated modules, while a three-phase schedule for colocated
modules prevents encoder collectives from conflicting with LLM pipeline P2P.
Together, boundary communicators, placement modes, and scheduling extensions
provide a runtime substrate for module-local parallelism in one end-to-end MLLM
training graph. We make the following contributions:

\begin{enumerate}
    \item \textbf{A heterogeneous-parallel runtime abstraction for MLLM
    training.} We let modules in one end-to-end multimodal graph choose
    independent TP/CP/PP/DP/EP layouts and physical rank placements, covering both
    colocated execution on shared GPUs and non-colocated execution on disjoint
    rank sets.

    \item \textbf{Boundary communicators and heterogeneous pipeline schedules.}
    We introduce autograd-aware boundary communicators that materialize forward
    activations in the destination layout and return backward gradients to the
    source layout. We compose these transforms with pipeline-parallel training
    using graph-aware 1F1B dispatch for non-colocated modules and a three-phase
    schedule for colocated modules.

    \item \textbf{An operating-regime evaluation against optimized homogeneous
    baselines.} We evaluate the two modes in the regimes they target:
    shared-GPU colocated workloads across encoder size, context length,
    vision-token density, and GPU scale, and workloads that pair larger encoders
    with a 120B LLM, where non-colocation separates encoder state from LLM ranks.
    Across these regimes, colocated execution
    improves TFLOPS/GPU\footnote{TFLOPS denotes tera floating-point operations
    per second; TFLOPS/GPU denotes per-GPU throughput.} by up to 49.3\%, while
    non-colocated execution improves aggregate token throughput by up to
    13.0\% and TFLOPS/GPU by up to 9.6\%.

    \item \textbf{Correctness, convergence, and open-source release.}
    Step-level FP32 parity checks and FP32/BF16\footnote{FP32 denotes
    single-precision 32-bit floating point; BF16 denotes bfloat16.} LLaVA-style
    training sweeps validate correctness and convergence, and we release the
    implementation in Megatron-LM for community use in large-scale multimodal
    training.
\end{enumerate}

%% file: tex/related.tex
\section{Related work}
\label{sec:related}

Large-scale training frameworks such as Megatron-LM, DeepSpeed/ZeRO,
Megatron-DeepSpeed, and PyTorch FSDP provide the core mechanisms for scaling
modern transformer training, including tensor, data, pipeline,
sequence/context, expert, and optimizer-state parallelism
~\citep{shoeybi2019megatron,narayanan2021megatron,
smith2022megatrondeepspeed,rajbhandari2020zero,zhao2023pytorchfsdp}.
These systems are highly effective for homogeneous transformer stacks, but
typically expose a single parallel layout for the composed graph. In multimodal
training, this shared-layout assumption can force encoders and LLMs into the
same TP/DP/PP/CP choices despite their different sequence lengths, memory
pressure, and communication patterns.

Several recent systems improve the efficiency of heterogeneous multimodal
training graphs without relaxing this shared-layout constraint. Prior work
reduces pipeline bubbles or wavefront overheads
~\citep{optimus2025,pipeweaver2025,graphpipe2024,spindle2025}, addresses
sequence or workload imbalance~\citep{hydraulis2024,flexsp2025,dcp2025,
dhp2025}, and mitigates multimodal data-plane bottlenecks
~\citep{orchmllm2025,overlord2025}. These techniques improve scheduling,
balancing, or data movement around a selected layout, but do not let adjacent
modules choose and compose independent parallel layouts inside one end-to-end
training graph.

More closely related systems explicitly target multimodal heterogeneity, but
their abstractions remain tied to particular topologies or placement models.
DISTMM~\citep{distmm2025} partitions heterogeneous multimodal submodules, but focuses on
contrastive or fusion-style models where modality towers meet at aggregation,
interaction, or loss modules. MLLM training instead has
feed-through boundaries: encoder activations become projector, embedding, and
LLM inputs, and gradients must return through those same edges. DistTrain~\citep{disttrain2025}
disaggregates the modality encoder, LLM backbone, and generator into
broker-connected parallelism units, but centers on
physically separated units rather than colocated modules that share GPUs while
using different logical layouts. Cornstarch exposes modular MLLM construction
and modality-aware parallelism~\citep{cornstarch2025,cornstarch_parallel_docs},
but its public interface still restricts independent module layouts, including
different encoder/LLM TP sizes, LLM-only CP, and DP inferred after composition.

In contrast, our work makes heterogeneous parallelism a first-class runtime
abstraction for MLLM training. Boundary communicators pair forward activation
materialization with backward gradient transforms, preserving tensor semantics
across independently parallelized modules. The same abstraction supports both
shared-rank colocated execution and disjoint-rank non-colocated execution, while
heterogeneous pipeline scheduling composes the modules into one end-to-end
training graph. By implementing this in Megatron-LM, we provide a reusable
runtime substrate for modular MLLM frameworks, disaggregated training systems,
and future placement planners.

%% file: tex/method.tex
\section{Method}
\label{sec:method}

\begin{figure*}[tbp]
  \centering
  \includegraphics[width=\textwidth]{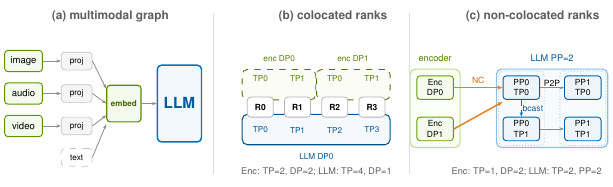}
  \caption{Multimodal graph and heterogeneous placement modes. Modality encoders produce projected tokens that are inserted into the LLM input sequence. Colocated execution reinterprets the same physical ranks under different encoder and LLM logical grids, while non-colocated execution places encoder and LLM on disjoint rank sets and uses non-colocated communicators before ordinary LLM pipeline P2P.}
  \label{fig:mimo-overview}
  \label{fig:deployment-modes}
  \label{fig:colocated-layout}
  \label{fig:non-colocated-layout}
\end{figure*}

We view multimodal training as a graph of modules and graph operators connected
by activation edges. As shown in Figure~\ref{fig:mimo-overview}(a), the graph
contains modality encoders, projection layers, an embedding assembly step, and
an autoregressive language model. Rather than assigning one parallel layout to
the entire graph, our method assigns each module \(m\) its own logical layout
\(G_m=(\mathrm{TP}_m,\mathrm{DP}_m,\mathrm{PP}_m,\mathrm{CP}_m,\ldots)\) and
physical rank set \(R_m\). The logical layout describes how the module shards or
replicates its computation; the rank set describes where that module executes.

A boundary edge \(u \rightarrow v\) carries one tensor with two distributed
representations. The source module produces it according to \(G_u\) on ranks
\(R_u\), while the destination module consumes it according to \(G_v\) on ranks
\(R_v\). A boundary communicator is the runtime primitive that converts between
these representations: forward materializes the activation in the destination
layout, and backward applies the matching reverse layout transform so that
gradients return to the source layout. This preserves boundary tensor semantics
while allowing adjacent modules to use independent TP/DP/PP/CP grids inside one
end-to-end training graph.

The boundary communicator implements a logical batch-shard transform for every inter-module edge.
Equal-DP boundaries pair source and destination shards one-to-one.
Fan-in, where $\text{DP}_u > \text{DP}_v$, maps multiple source batch intervals to one destination shard; backward splits the destination gradient along the same recorded intervals.
Fan-out, where $\text{DP}_u < \text{DP}_v$, maps one source shard to multiple destination intervals; backward concatenates sibling destination-gradient intervals to reconstruct the source-layout gradient.
How this logical transform is physically realized depends on whether source and destination modules share ranks or occupy disjoint rank sets; we therefore introduce colocated and non-colocated communicators next.

\subsection{Non-colocated communicator}
\label{sec:method-non-colocated}

A non-colocated boundary realizes the layout transform through explicit transfer
between the source module's last pipeline stage and the destination module's
first pipeline stage. The performance-sensitive step is the cross-boundary
exchange: module-local TP/CP groups are often mapped within a high-bandwidth
locality domain, while disjoint module islands may communicate over
lower-bandwidth fabric. To avoid a cross-product exchange between source and
destination TP/CP ranks, the communicator routes each boundary DP shard through a
deterministic boundary leader, usually the rank with canonical TP/CP coordinates
on the boundary stage. Leaders perform all inter-module P2P; local TP/CP
broadcast then materializes the activation or gradient on the non-leader ranks
required by the receiving layout.

This design makes cross-boundary communication follow the logical batch-shard
relation rather than the product of source and destination TP/CP widths. The
source and destination leaders are derived from \(G_u\), \(G_v\), \(R_u\), and
\(R_v\), so both sides construct the same static routing pattern for the edge.
Equal-DP boundaries use one-to-one P2P between paired source and destination
leaders. Fan-in uses a many-to-one route: source leaders whose batch intervals
form one destination shard send to the same destination leader, which
concatenates in batch order. Fan-out uses the dual one-to-many route: a source
leader partitions its shard into destination intervals and sends each interval to
the corresponding destination leader.

\begin{figure*}[tbp]
  \centering
  \includegraphics[width=\textwidth]{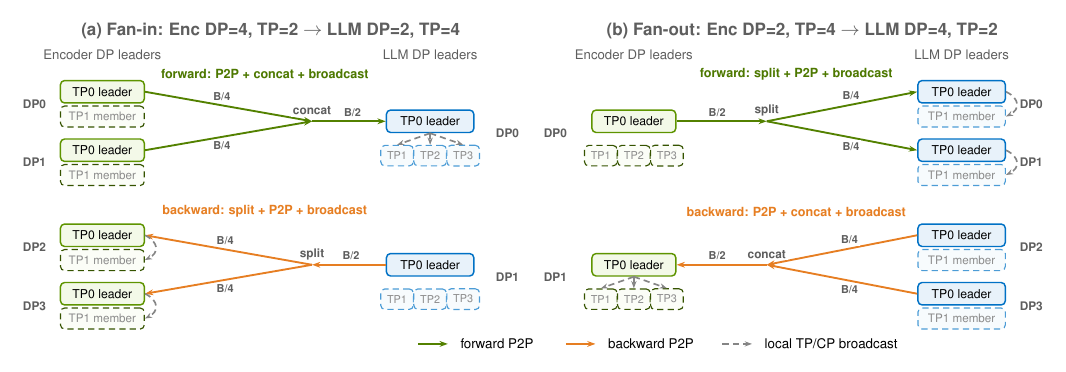}
  \caption{Non-colocated boundary communicator. Source and destination modules occupy disjoint rank sets, so boundary tensors cross the module boundary only through deterministic boundary leaders. Solid arrows show inter-module P2P among leaders; dashed arrows show local TP/CP broadcast to non-leader ranks. \(B\) denotes the global boundary batch. In fan-in, two \(B/4\) source shards are concatenated into one \(B/2\) destination shard in forward, and the \(B/2\) gradient is split back into two \(B/4\) source gradients in backward. Fan-out is the dual case: one \(B/2\) source shard is split into two \(B/4\) destination shards in forward, and the returned gradients are concatenated in backward. Unequal TP widths change the local broadcast domains around each leader, not the number of cross-boundary rank pairs.}
  \label{fig:noncolocated-communicator}
\end{figure*}

Backward uses the inverse leader routes. For equal-DP, the destination leader
returns the gradient shard to the paired source leader. For fan-in, the
destination leader splits the gradient along the recorded batch intervals and
returns each slice to the source leader that supplied the corresponding forward
shard. For fan-out, the source leader receives sibling destination-gradient
intervals and concatenates them to reconstruct the source-layout gradient. In
all cases, the reconstructed leader result is broadcast within the local TP/CP
group required by the receiving side. Figure~\ref{fig:noncolocated-communicator}
visualizes these leader routes for unequal TP widths.

\subsection{Colocated communicator}
\label{sec:method-colocated}

A colocated boundary realizes the same layout transform on a shared rank set.
The source and destination modules use the same physical ranks but assign each
rank different logical coordinates under \(G_u\) and \(G_v\). The communicator
is an autograd-aware shared-rank layout operator: its forward pass constructs the
destination-layout representation on the shared ranks, and its backward pass
reconstructs the corresponding source-layout gradient representation.

For each boundary edge, the runtime derives every rank's source coordinate,
destination coordinate, batch interval, and local collective group from \(G_u\),
\(G_v\), and the shared rank set. Equal-DP boundaries are rank-local: each
destination rank consumes the source shard already present on that physical rank,
reinterpreted under the destination layout. Fan-in uses local all-gather among
ranks whose source intervals form one destination shard, followed in backward by
interval selection from the destination-gradient shard. Fan-out is the dual:
forward selects the destination batch interval from the larger source shard
locally, while backward all-gathers sibling destination-gradient intervals to
reconstruct the original source-layout gradient.

\begin{figure*}[tbp]
  \centering
  \includegraphics[width=\textwidth]{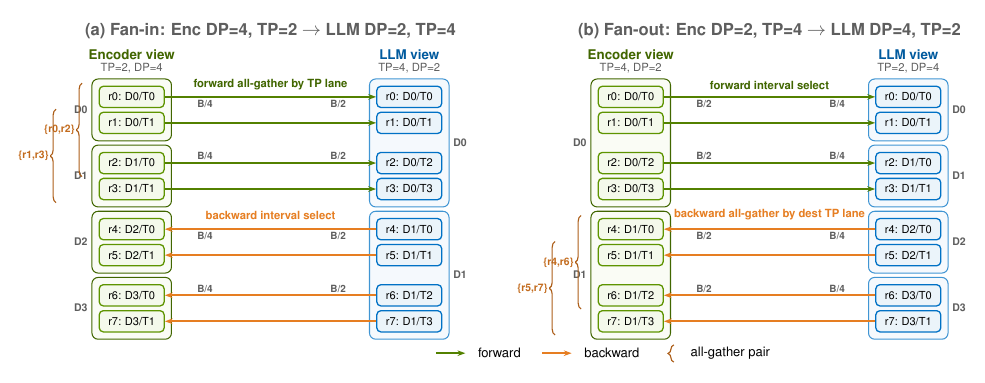}
  \caption{Colocated boundary communicator. Source and destination modules share the same physical ranks but interpret them under different logical layouts. \(B\) denotes the global boundary batch. In fan-in, ranks whose source slices form one destination shard all-gather locally: two \(B/4\) source slices become one \(B/2\) destination shard in forward, and backward selects from the \(B/2\) gradient the batch interval contributed by each source rank. In fan-out, forward is rank-local batch-interval selection from a \(B/2\) source shard to a \(B/4\) destination shard, while backward all-gathers sibling \(B/4\) gradients to reconstruct the original \(B/2\) source-gradient shard. Thus, colocated communication changes the logical view of a shared rank set rather than moving tensors across rank sets.}
  \label{fig:colocated-communicator}
\end{figure*}

Figure~\ref{fig:colocated-communicator} visualizes these shared-rank transforms.
Unlike the non-colocated protocol, colocated communication does not move tensors
across rank sets; it changes the logical view of the shared rank set and invokes
local collectives only when the layout transform spans multiple batch intervals.
We profile these communicators and find that their overhead is less than 1\%;
Appendix~\ref{app:bridge-communication} provides details.
Section~\ref{sec:method-schedules} describes how this shared-rank layout
operator is ordered with the LLM pipeline schedule.

\subsection{Pipeline orchestration over heterogeneous modules}
\label{sec:method-schedules}
\label{sec:method-noncolocated-schedule}

\begin{figure*}[tbp]
  \centering
  \includegraphics[width=\textwidth]{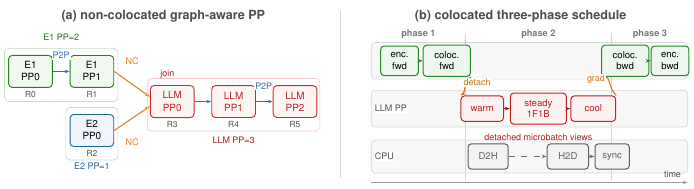}
  \caption{Pipeline orchestration for heterogeneous modules. (a) Non-colocated graph-aware pipeline dispatch: Encoder~1 has two PP stages, Encoder~2 has one, and the LLM has three; intra-module edges use module-local P2P, while encoder--LLM edges invoke edge-local non-colocated communicators. (b) Colocated LLM-PP training uses three phases: encoder forward and colocated forward transform, detached LLM 1F1B, then boundary-gradient handoff followed by colocated backward transform and encoder backward.}
  \label{fig:pipeline-orchestration}
  \label{fig:noncolocated-dag-topology}
  \label{fig:colocated-three-phase-schedule}
\end{figure*}

The boundary communicators define the layout transform for one edge; the
scheduler decides when those edge transforms are invoked inside the full
training loop. This is not ordinary linear pipeline parallelism. A conventional
1F1B schedule assumes one linear PP chain, where each stage communicates with a
fixed predecessor and successor. In a heterogeneous multimodal graph, modules
can have different PP depths and layouts, a node can have multiple incoming or
outgoing activation edges, and a rank's next communication may be either
module-local P2P or a boundary communicator that changes layouts across modules.
A pipeline action is therefore an edge-dispatch decision over a PP-stage graph,
not simply ``send to the next PP rank.''

For non-colocated execution, the runtime expands the module graph into PP-stage
nodes, as in Figure~\ref{fig:pipeline-orchestration}(a). At each schedule action
and microbatch, it resolves the active module edge. Internal edges use the
module's ordinary PP communicator; edges crossing disjoint module rank sets
invoke the non-colocated communicator from
Section~\ref{sec:method-non-colocated} between the source module's final PP
stage and the destination module's first PP stage. Preserving edge identity is
essential at joins and branches: an LLM join becomes runnable only after all
required encoder activations have been materialized in the LLM layout, and
backward routes each boundary gradient through the communicator attached to the
forward edge that produced that activation.

The scheduler also generalizes warmup, steady state, and cooldown. In a linear
pipeline, these phases are derived from a stage's position in one chain. In a
PP-stage graph, different nodes can have different downstream distances to the
sink, and these distances determine when each node warms up, enters paired
forward/backward actions, and drains. At steady state, the two halves of a
paired schedule call may resolve to different edge types: a rank can send a new
activation through one edge while receiving an older gradient through the reverse
of another. Appendix~\ref{app:pipeline-schedules} expands
Figure~\ref{fig:pipeline-orchestration}(a) into the corresponding dispatch
table.

Colocated execution introduces a different ordering constraint. The encoder and
LLM share physical ranks, but their communication order is not the same. Encoder
execution may require collectives over the encoder layout, where all ranks in a
collective group must enter together. LLM 1F1B deliberately staggers ranks
across PP stages and microbatches. If encoder collectives are invoked inside the
LLM pipeline loop, ranks in the same encoder collective group can arrive at
different times; adding global synchronization would avoid the conflict but
serialize the LLM pipeline.

We therefore use the three-phase colocated schedule in
Figure~\ref{fig:pipeline-orchestration}(b). First, the runtime batches the
samples assigned to the colocated rank group for the current LLM pipeline
window, runs encoder forward once over those microbatches, and applies the
colocated forward transform from Section~\ref{sec:method-colocated} to produce a
packed boundary tensor in the LLM PP0 layout. Second, this tensor is detached as
a leaf boundary variable and exposed to LLM PP0 as per-microbatch views;
ordinary LLM 1F1B then indexes the active view, inserts it into that microbatch's
text embeddings, and proceeds with standard LLM pipeline P2P without re-entering
encoder collectives. Third, after the LLM pipeline drains, accumulated gradients
on the packed boundary tensor are handed through the colocated backward
transform, and encoder backward runs over the saved batched encoder activations.

The explicit gradient handoff reconnects the detached LLM phase to the
encoder-side autograd graph while preserving the paired forward/backward layout
transform. The phase boundary also creates a memory window: encoder parameters
can be offloaded after encoder forward, reloaded during LLM cooldown, and
synchronized before encoder backward consumes them.

\FloatBarrier

%% file: tex/experiments.tex
\section{Experimental results}
\label{sec:experiments}

We evaluate how independent encoder and LLM layouts improve end-to-end MLLM
training throughput. A homogeneous shared TP/CP/PP/DP grid imposes one set
of layout choices on both encoder and LLM, which can over-constrain the system.
Heterogeneous parallelism relaxes this constraint by allowing each module to use
its own layout. This flexibility matters because the optimization space is
large: encoder scale, context length, vision-token density, world size, and
optimizer-state sharding each shift pressure among compute, memory, pipeline
balance, and communication. As these pressures shift, the best operating point
can also change, including whether heterogeneity should be colocated or
non-colocated. Our evaluation samples representative points across this
landscape to characterize where each execution mode is most effective.

For each workload, we tune homogeneous (H), colocated heterogeneous
(\textbf{C}), and non-colocated heterogeneous (NC) execution to their best valid
throughput configurations.\footnote{Throughput is measured as the median over
15 steady-state iterations after warmup; repeated measurements on representative
configurations showed low run-to-run variation, typically below 1\%.} The search includes module layouts, recomputation,
batch-size constraints, distributed optimizer instances, and LLM pipeline-layer
balance. Colocated runs use the same physical world size and global batch size
as homogeneous training, whereas non-colocated runs are iso-GBS\footnote{GBS
denotes global batch size.}; when exact world-size matching is infeasible, we
keep the LLM world size comparable to homogeneous training, add a compact
encoder island, and report both aggregate tokens/s and TFLOPS/GPU.
Appendix~\ref{app:opensource-config} shows how these module layouts and
placements are expressed in the user-facing configuration.

\begin{figure*}[!t]
  \centering
  \includegraphics[width=\textwidth]{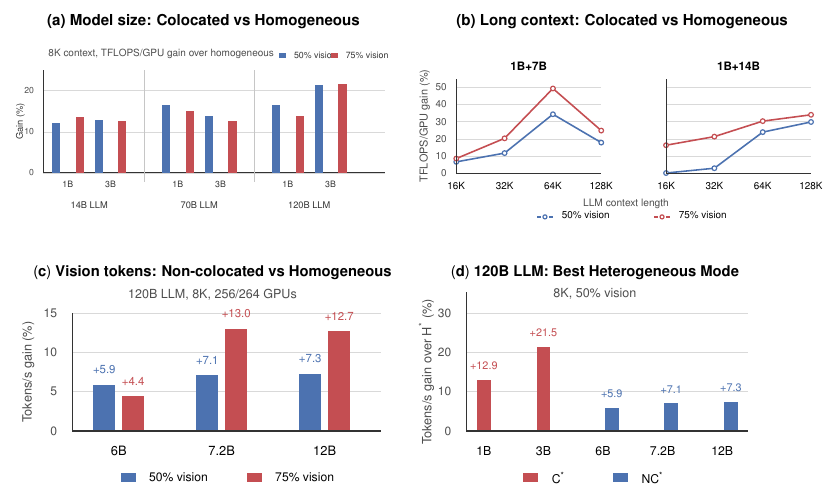}
  \caption{Operating regimes for heterogeneous training. When memory headroom is
  sufficient, colocated execution is the right first step: it gives the encoder
  and LLM separate logical layouts while keeping them on the same GPUs. Long
  context strengthens this effect because the LLM needs context parallelism
  while the encoder does not; keeping encoder \(\mathrm{CP}=1\) turns LLM
  context-parallel ranks into encoder data-parallel replicas. As encoder
  pressure grows with a 120B LLM, non-colocated execution becomes the better
  operating point: it separates encoder work from the LLM rank set, and the best
  heterogeneous mode switches from colocated for small encoders to non-colocated
  from the 6B encoder scale onward. Appendices~\ref{app:colocated-throughput-results},
  \ref{app:noncolocated-throughput-results}, and~\ref{app:operating-ranges}
  report the full measurements.}
  \label{fig:throughput-summary}
\end{figure*}

We first evaluate colocated heterogeneous execution, which is the relevant
setting when the encoder and LLM can fit on the same GPUs but should not share
the same logical grid. In this regime, heterogeneous parallelism improves
throughput by remapping the encoder and LLM to different layouts. The encoder
can reduce unnecessary TP, avoid unnecessary CP, increase useful DP, and choose
its own number of distributed optimizer instances. This last choice is important
because increasing encoder DP can improve encoder efficiency, but it also
changes how optimizer state is sharded and how much optimizer communication is
introduced. Colocated heterogeneity lets the encoder make this tradeoff
independently of the LLM.

Figure~\ref{fig:throughput-summary}(a) summarizes this shared-GPU regime at 8K
context, with full measurements in Table~\ref{tab:app-colocated-raw}. Across the
short-context model-size sweep, colocated execution consistently improves
throughput when the shared homogeneous grid over-constrains the encoder or the
LLM pipeline. The tuned 256-GPU rows show the same mechanism at larger scale,
where encoder layout and encoder optimizer instances are selected per workload.

As an example, for the 1B encoder + 14B LLM at 8K, 50\% vision tokens, and 72
GPUs, the best homogeneous configuration uses a shared 4/1/1/18 TP/CP/PP/DP
layout. The best colocated heterogeneous configuration instead uses encoder
2/1/1/36 and LLM 2/1/2/18. This lets the LLM use PP=2 with lower TP, while the
encoder remains a single-stage, higher-DP module rather than being tied to the
LLM pipeline structure. Step time improves at the same GPU count and essentially
the same peak memory usage.

As context length grows, the LLM needs CP to fit and parallelize the fused
multimodal sequence, but the encoder sequence length is fixed by image
resolution and patching; colocated heterogeneity keeps encoder CP=1 and converts
those ranks into encoder DP replicas. This is the strongest colocated setting
in our sweep (Figure~\ref{fig:throughput-summary}(b) and
Table~\ref{tab:app-long-context-raw}).

We next evaluate non-colocated heterogeneous execution, which targets a
different bottleneck. For larger encoders paired with the 120B LLM, the active
constraint shifts from logical layout coupling to memory residency. Encoder
parameters, activations, and optimizer state now compete with the LLM for the
same rank memory and can block a better LLM pipeline layout. In this regime,
non-colocated heterogeneity becomes the better operating point: the encoder runs
on a compact island, while the LLM recovers an independently optimized layout
over the remaining ranks.

Figure~\ref{fig:throughput-summary}(c) and (d) show how the best operating
point for a 120B LLM changes between colocated and non-colocated modes, with
full measurements in Tables~\ref{tab:app-noncolocated-raw}
and~\ref{tab:app-operating-ranges}. For 1B and 3B encoders, colocated execution
remains the better operating point because the encoder can share the LLM GPUs
and benefit from layout remapping. Starting at the 6B encoder scale, the best
mode switches to non-colocated execution. The NC rows remain positive despite
averaging over the additional 8-GPU encoder island. This indicates that, once
encoder residency constrains the LLM memory or pipeline layout, separating the
encoder can improve both end-to-end throughput and normalized per-GPU
efficiency.

Overall, the experiments show that heterogeneous parallelism improves MLLM
throughput by making module-level layout and placement tunable. Colocated
execution extracts more throughput from a fixed GPU budget when encoder and LLM
can share hardware but need different logical grids. Non-colocated execution
becomes effective when encoder residency should be separated from the LLM rank
set. This flexibility lets the system select the right operating point for each
workload.

\FloatBarrier

%% file: tex/convergence_validation.tex
\section{Convergence validation}
\label{sec:convergence-validation}

Heterogeneous parallelism should change only how tensors are partitioned, placed,
and communicated across ranks, not the underlying training semantics. Because
training loss curves can hide subtle communicator-level errors, we first validated
this invariant with step-level numerical parity tests in FP32. Starting from the
same initialization and identical inputs, the homogeneous baseline, a
non-colocated variant, and a colocated heterogeneous variant execute identical
training steps; after each optimizer update, we compare all parameters,
gradients, and optimizer states within tight numerical tolerance. This directly
checks boundary-communicator correctness, including forward activation
materialization and backward gradient routing.

For convergence parity, we compare loss trajectories across colocated and non-colocated layouts using a LLaVA-1.5-style workload: a CLIP ViT-L/14 vision encoder~\citep{radford2021learning}, a two-layer projection MLP, and Vicuna-7B~\citep{vicuna2023} trained on LLaVA-Pretrain. LLaVA serves as a representative validation case, but our boundary communication mechanism also supports cross-attention-based MLLM such as Qwen3VL. We use FP32 (vision encoder frozen, projection MLP and LLM trainable), disable TF32, and enforce deterministic PyTorch/cuDNN execution and fixed NCCL collective selection to minimize numerical noise. Because TP reductions remain layout-order-dependent, we compare within numerical tolerance rather than requiring bitwise agreement. Appendix~\ref{app:llava-setup} provides the full validation protocol and extends the study to longer BF16 runs and broader TP/PP/DP layout sweeps.

Figure~\ref{fig:convergence-homo-hetero} reports the main controlled convergence
comparison. Starting from the homogeneous baseline, the same-layout
non-colocated variant changes physical placement while keeping module layouts
fixed, and the colocated heterogeneous variant changes module-local layout while
keeping placement fixed. All other training settings are held constant. The loss
curves closely overlap, supporting convergence parity under both placement and
layout changes.

%% file: tex/appendix.tex
\clearpage
\onecolumn
\appendix
\setcounter{table}{0}
\renewcommand{\thetable}{A\arabic{table}}
\renewcommand{\theHtable}{appendix.table.A\arabic{table}}
\input{tex/appendix_throughput_tables.tex}

\section{Open-source user configuration example}
\label{app:opensource-config}

\lstdefinestyle{mimoconfig}{
  language=Python,
  basicstyle=\ttfamily\scriptsize,
  keywordstyle=\color{RoyalBlue}\bfseries,
  stringstyle=\color{ForestGreen!65!black},
  commentstyle=\color{black!55},
  showstringspaces=false,
  columns=fullflexible,
  keepspaces=true,
  xleftmargin=0pt,
  xrightmargin=0pt,
  alsoletter={_},
  morekeywords={
    parallelism,
    module_parallelisms,
    tensor_model_parallel_size,
    pipeline_model_parallel_size,
    data_parallel_size,
    rank_offset
  },
  emph={MegatronMIMOParallelismConfig,ModuleParallelismConfig},
  emphstyle=\color{BurntOrange!90!black}\bfseries
}

\begin{center}
\begin{minipage}[t]{0.48\linewidth}
\scriptsize
\textbf{Non-colocated: disjoint GPU sets}
\vspace{0.25em}
\begin{lstlisting}[style=mimoconfig]
...
parallelism = MegatronMIMOParallelismConfig(
    module_parallelisms={
        "language": ModuleParallelismConfig(
            tensor_model_parallel_size=2,
            pipeline_model_parallel_size=2,
            data_parallel_size=1,
            rank_offset=0,   # ranks [0, 4)
        ),
        "images": ModuleParallelismConfig(
            tensor_model_parallel_size=1,
            pipeline_model_parallel_size=1,
            data_parallel_size=4,
            rank_offset=4,   # ranks [4, 8)
        ),
    },
)
...
\end{lstlisting}
\end{minipage}
\hfill
\begin{minipage}[t]{0.48\linewidth}
\scriptsize
\textbf{Colocated: shared GPU set}
\vspace{0.25em}
\begin{lstlisting}[style=mimoconfig]
...
parallelism = MegatronMIMOParallelismConfig(
    module_parallelisms={
        "language": ModuleParallelismConfig(
            tensor_model_parallel_size=4,
            pipeline_model_parallel_size=1,
            data_parallel_size=2,
            rank_offset=0,   # ranks [0, 8)
        ),
        "images": ModuleParallelismConfig(
            tensor_model_parallel_size=1,
            pipeline_model_parallel_size=1,
            data_parallel_size=8,
            rank_offset=0,   # ranks [0, 8)
        ),
    },
)
...
\end{lstlisting}
\end{minipage}
\captionsetup{hypcap=false}
\captionof{figure}{Minimal user-facing configuration for heterogeneous
parallelism in multimodal training. The left example places the language model
and vision encoder on disjoint rank ranges, so the runtime uses the
non-colocated communicator. The right example places both modules on the same
physical rank range, so the runtime uses colocated execution while still
allowing different module-local TP/PP/DP layouts.}
\label{fig:opensource-config}
\end{center}

The open-source implementation exposes this layout choice through a compact
configuration interface. Users provide one entry per module, specifying its
TP/PP/DP layout and rank offset; together, these fields determine the module's
logical process groups and physical rank range. At model construction time, the
runtime uses the resulting ranges to select colocated or non-colocated boundary
communicators, while keeping the model definition, loss, and training loop
shared across both placement modes.

\section{LLaVA validation details and additional convergence results}
\label{app:llava-setup}

\subsection{Validation setup}
\label{app:llava-validation-setup}

This appendix gives the convergence-validation details that complement
Section~\ref{sec:convergence-validation}. Appendix~\ref{app:llava-validation-setup}
describes the LLaVA setup and deterministic FP32 validation protocol,
Appendix~\ref{app:additional-convergence} reports the additional frozen-LLM and
BF16 convergence results, and Appendices~\ref{app:convergence-config-sweep}
and~\ref{app:convergence-colocated} give the trainable-LLM non-colocated and
colocated layout sweeps.

The validation model follows LLaVA-1.5. 
We use a CLIP ViT-L/14 vision encoder, a two-layer projection MLP, and a Vicuna-7B v1.5 language model based on Llama-2-7B. 
The vision encoder and language model are initialized from pretrained CLIP ViT-L/14 and Vicuna-7B checkpoints, respectively. 
Across all validation runs, the vision encoder is frozen and the projection MLP is trainable; the LLM is either frozen for projector-only runs or trainable for trainable-LLM runs.
The projection MLP initializes the first linear layer weights from \(\mathcal{N}(0,0.02)\) with zero bias; the second linear layer weights are initialized from \(\mathcal{N}(0,0.02/\sqrt{2})\), also with zero bias.

The vision tower has 24 layers and 16 attention heads, processes \(336 \times 336\) images with \(14 \times 14\) patches, and contributes 576 patch features of width 1024 after dropping the CLS token.
The projection MLP maps these features to the 4096-dimensional language-model hidden space.
Vicuna-7B v1.5 uses 32 decoder layers, 32 attention heads, hidden size 4096, RoPE base 10,000, and a 4096-token context length.
For \(\mathrm{TP}=4\), Megatron-LM pads the Vicuna vocabulary plus the reserved LLaVA image token to 32,256 entries.

For the FP32 parity runs, we further reduce numerical nondeterminism so loss
curves can be compared more directly across layouts. We switch these runs to
FP32 precision, disable TF32, and enable PyTorch deterministic execution with
\texttt{torch.\allowbreak use\_\allowbreak deterministic\_\allowbreak algorithms(True)}. We also
disable cuDNN autotuning with \texttt{cudnn.benchmark=False} and set
\texttt{cudnn.deterministic=True} for deterministic cuDNN kernels. For NCCL
reductions, we pin the algorithm to Ring so collective reductions use a
deterministic order across runs. These settings reduce loss-curve variation
caused by floating-point non-associativity, making discrepancies more likely to
reflect communicator errors rather than runtime nondeterminism.

\begin{table}[!ht]
  \centering
  \small
  \caption{Optimization hyperparameters for the LLaVA convergence validation.}
  \label{tab:convergence-training}
  \begin{tabularx}{\columnwidth}{@{}>{\raggedright\arraybackslash}p{0.38\columnwidth}X@{}}
    \toprule
    Parameter & Value \\
    \midrule
    Global batch size & 96 \\
    Optimizer & Distributed Adam \\
    Learning rate & 1.0e-3 (1.0e-4 for trainable LLM) \\
    Min learning rate & 2.0e-5 (1.0e-5 for trainable LLM) \\
    Schedule / warmup & Cosine decay / 60 steps \\
    Weight decay & 0.0 \\
    \(\beta_1/\beta_2\) & 0.9 / 0.95 \\
    Gradient clipping & 0.0 (1.0 for BF16 runs) \\
    \bottomrule
  \end{tabularx}
\end{table}

\subsection{Additional convergence validation results}
\label{app:additional-convergence}

\begin{figure}[!t]
  \centering
  \includegraphics[width=\linewidth]{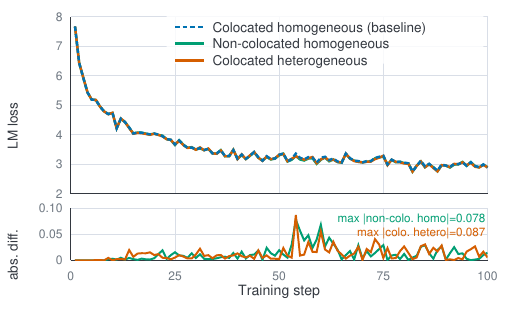}
  \caption{Frozen-LLM FP32 placement and layout convergence comparison. The
  vision encoder and LLM are frozen, so only the projection MLP is trained. The
  modified runs follow the corresponding colocated homogeneous baseline.}
  \label{fig:app-convergence-frozen-homo-hetero}
  \label{fig:app-convergence-additional-homo-hetero}
\end{figure}

\begin{figure}[!t]
  \centering
  \begin{subfigure}[t]{0.49\linewidth}
    \centering
    \includegraphics[width=\linewidth]{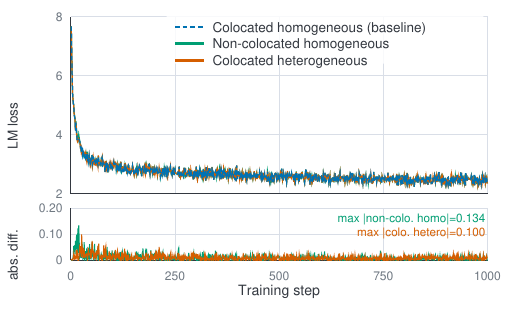}
    \caption{Frozen LLM, BF16.}
    \label{fig:app-convergence-frozen-homo-hetero-bf16}
  \end{subfigure}
  \hfill
  \begin{subfigure}[t]{0.49\linewidth}
    \centering
    \includegraphics[width=\linewidth]{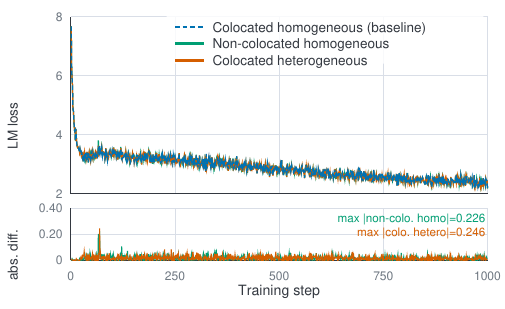}
    \caption{Trainable LLM, BF16.}
    \label{fig:app-convergence-trainable-homo-hetero-bf16}
  \end{subfigure}
  \caption{BF16 placement and layout convergence comparisons. Panel (a) uses
  projector-only training with the vision encoder and LLM frozen; panel (b)
  keeps the vision encoder frozen while training the LLM and projection MLP. In
  both cases, the modified runs follow the corresponding colocated homogeneous
  baseline.}
  \label{fig:app-convergence-bf16-homo-hetero}
\end{figure}

\begin{figure}[!t]
  \centering
  \begin{subfigure}[t]{0.49\linewidth}
    \centering
    \includegraphics[width=\linewidth]{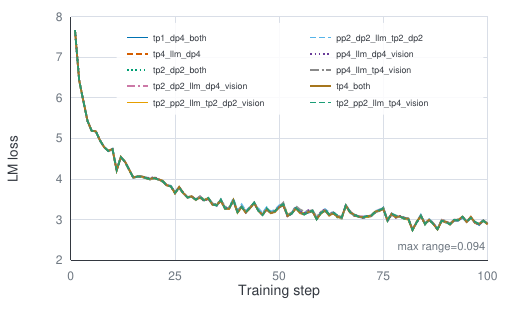}
    \caption{Non-colocated frozen-LLM layout sweep.}
    \label{fig:app-convergence-frozen-hetero-sweep}
  \end{subfigure}
  \hfill
  \begin{subfigure}[t]{0.49\linewidth}
    \centering
    \includegraphics[width=\linewidth]{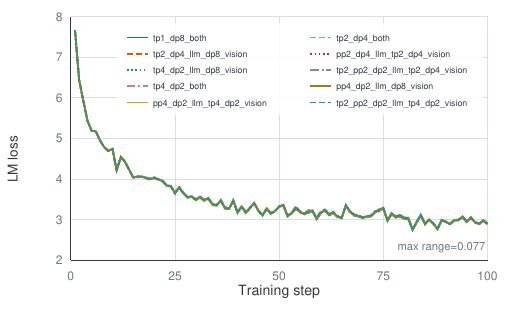}
    \caption{Colocated frozen-LLM layout sweep.}
    \label{fig:app-convergence-colocated-frozen-sweep}
  \end{subfigure}
  \caption{Frozen-LLM configuration sweeps. Panel (a) sweeps the non-colocated
  configurations from Table~\ref{tab:convergence-frozen-configs}; panel (b)
  applies the corresponding colocated layouts by sharing the physical ranks and
  doubling each module's DP degree. The overlapping curves show that convergence
  remains insensitive to the tested parallelism layout when only the projection
  MLP is trained.}
  \label{fig:app-convergence-frozen-sweeps}
\end{figure}

Section~\ref{sec:convergence-validation} and
Appendices~\ref{app:convergence-config-sweep}--\ref{app:convergence-colocated}
focus on FP32 trainable-LLM runs.
Here we report the remaining convergence validation results: the FP32
projector-only frozen-LLM placement/layout comparison, the BF16
placement/layout comparisons, and the frozen-LLM layout sweeps.
The shorter validation runs use 100 training steps in FP32 precision, whereas
the larger-scale comparisons run for 1K training steps in mixed-precision BF16.
Runs within each experiment set use the same initialization, data order, and
optimization setup, with hyperparameters summarized in
Table~\ref{tab:convergence-training}. The frozen-LLM layout sweep uses the
configurations in Table~\ref{tab:convergence-frozen-configs}.
Figure~\ref{fig:app-convergence-frozen-homo-hetero} shows the FP32
projector-only placement/layout comparison, Figure~\ref{fig:app-convergence-bf16-homo-hetero}
reports the corresponding BF16 placement/layout comparisons, and
Figure~\ref{fig:app-convergence-frozen-sweeps} summarizes the frozen-LLM layout
sweeps.

\begin{table}[!ht]
  \centering
  \scriptsize
  \setlength{\tabcolsep}{1pt}
  \renewcommand{\arraystretch}{0.92}
  \begin{minipage}[t]{0.49\linewidth}
    \centering
    \captionof{table}{Non-colocated parallelism sweep with frozen language
    model. All configurations use 8 GPUs.}
    \label{tab:convergence-frozen-configs}
    \begin{tabularx}{\linewidth}{@{}>{\raggedright\arraybackslash}Xcc@{}}
      \toprule
      Config name & \begin{tabular}[c]{@{}c@{}}LLM\\[-1pt](TP/PP/DP)\end{tabular} &
      \begin{tabular}[c]{@{}c@{}}Vision\\[-1pt](TP/PP/DP)\end{tabular} \\
      \midrule
      \path{tp4_both} & 4 / 1 / 1 & 4 / 1 / 1 \\
      \path{tp2_dp2_both} & 2 / 1 / 2 & 2 / 1 / 2 \\
      \path{tp2_pp2_llm_tp4_vision} & 2 / 2 / 1 & 4 / 1 / 1 \\
      \path{tp1_dp4_both} & 1 / 1 / 4 & 1 / 1 / 4 \\
      \path{pp4_llm_tp4_vision} & 1 / 4 / 1 & 4 / 1 / 1 \\
      \path{pp4_llm_dp4_vision} & 1 / 4 / 1 & 1 / 1 / 4 \\
      \path{pp2_dp2_llm_tp2_dp2} & 1 / 2 / 2 & 2 / 1 / 2 \\
      \path{tp4_llm_dp4_vision} & 4 / 1 / 1 & 1 / 1 / 4 \\
      \path{tp2_pp2_llm_tp2_dp2_vis} & 2 / 2 / 1 & 2 / 1 / 2 \\
      \path{tp2_dp2_llm_dp4_vision} & 2 / 1 / 2 & 1 / 1 / 4 \\
      \bottomrule
    \end{tabularx}
  \end{minipage}
  \hfill
  \begin{minipage}[t]{0.49\linewidth}
    \centering
    \captionof{table}{Non-colocated parallelism sweep with trainable language
    model. All configurations use 8 GPUs.}
    \label{tab:convergence-unfrozen-configs}
    \begin{tabularx}{\linewidth}{@{}>{\raggedright\arraybackslash}Xcc@{}}
      \toprule
      Config name & \begin{tabular}[c]{@{}c@{}}LLM\\[-1pt](TP/PP/DP)\end{tabular} &
      \begin{tabular}[c]{@{}c@{}}Vision\\[-1pt](TP/PP/DP)\end{tabular} \\
      \midrule
      \path{tp4_both} & 4 / 1 / 1 & 4 / 1 / 1 \\
      \path{tp2_dp2_both} & 2 / 1 / 2 & 2 / 1 / 2 \\
      \path{tp2_pp2_llm_tp4_vision} & 2 / 2 / 1 & 4 / 1 / 1 \\
      \path{tp2_pp2_llm_tp2_dp2_vis} & 2 / 2 / 1 & 2 / 1 / 2 \\
      \path{pp4_llm_tp4_vision} & 1 / 4 / 1 & 4 / 1 / 1 \\
      \path{pp4_llm_tp2dp2_vision} & 1 / 4 / 1 & 2 / 1 / 2 \\
      \path{pp2_dp2_llm_tp2dp2_vis} & 1 / 2 / 2 & 2 / 1 / 2 \\
      \path{tp4_llm_tp2dp2_vision} & 4 / 1 / 1 & 2 / 1 / 2 \\
      \bottomrule
    \end{tabularx}
  \end{minipage}
\end{table}

\subsection{Non-colocated layout sweep}
\label{app:convergence-config-sweep}
\label{sec:convergence-config-sweep}

The second experiment set sweeps TP/PP/DP choices for non-colocated layouts
across the language model and vision encoder. These configurations preserve the
same end-to-end LLaVA training task while changing the ownership layout
presented to the boundary communicator. Each non-colocated sweep configuration
assigns four GPUs to the LLM and four GPUs to the vision encoder; the listed
TP/PP/DP triples show how each module partitions its own four-rank group.
Table~\ref{tab:convergence-unfrozen-configs} lists the trainable-LLM sweep,
which uses micro batch size (MBS) 2.
Figure~\ref{fig:convergence-trainable-sweeps}(a) shows that the loss curves
remain aligned across these non-colocated layouts when the LLM and projection
MLP are trainable.

\subsection{Colocated layout sweep}
\label{app:convergence-colocated}
\label{sec:convergence-colocated}

The third experiment set validates the colocated communicator under the same
trainable-LLM layout patterns tested in the non-colocated sweep. We reuse the
configurations from Table~\ref{tab:convergence-unfrozen-configs}, but run the
vision encoder and language model on the same eight physical ranks. Because each
module now spans the shared 8-GPU rank set rather than its own 4-GPU rank set,
we keep each module's TP and PP degrees unchanged and double its DP degree.
Thus, for example, a non-colocated configuration with
\(\mathrm{TP}=4,\mathrm{PP}=1,\mathrm{DP}=1\) for a module becomes
\(\mathrm{TP}=4,\mathrm{PP}=1,\mathrm{DP}=2\) in the colocated sweep.

\begin{figure}[!t]
  \centering
  \begin{subfigure}[t]{0.49\linewidth}
    \centering
    \includegraphics[width=\linewidth]{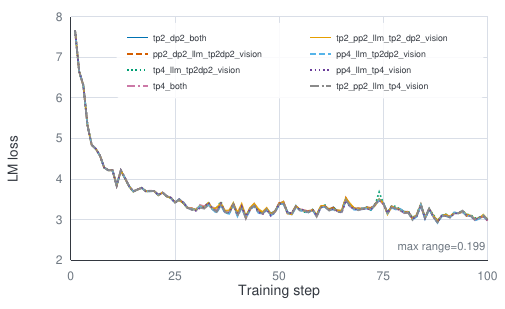}
    \caption{Non-colocated layout sweep.}
    \label{fig:convergence-hetero-sweep}
  \end{subfigure}
  \hfill
  \begin{subfigure}[t]{0.49\linewidth}
    \centering
    \includegraphics[width=\linewidth]{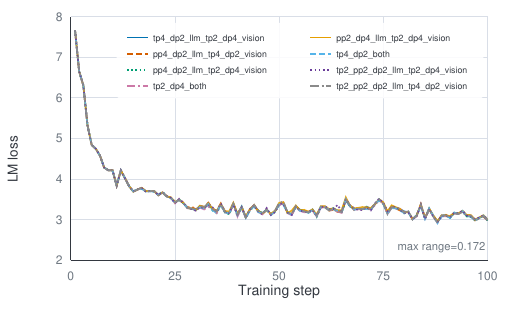}
    \caption{Colocated layout sweep.}
    \label{fig:convergence-colocated-sweep}
  \end{subfigure}
  \caption{Trainable-LLM FP32 configuration sweeps. The vision encoder is
  frozen, while the LLM and projection MLP are trained. Panel (a) sweeps
  non-colocated TP/PP/DP configurations from
  Table~\ref{tab:convergence-unfrozen-configs}; panel (b) sweeps the
  corresponding colocated configurations by sharing the physical ranks and
  doubling each module's DP degree. The overlapping trends indicate that
  convergence is insensitive to the tested parallelism layout. The annotations
  report the maximum step-wise loss range across all plotted configurations.}
  \label{fig:convergence-trainable-sweeps}
\end{figure}

Figure~\ref{fig:convergence-trainable-sweeps}(b) shows that the colocated loss
curves remain tightly aligned across the tested trainable-LLM configurations,
indicating that colocated rank reinterpretation and the corresponding
forward/backward ownership transforms preserve the LLaVA training signal across
the tested TP/PP/DP layouts.

\FloatBarrier

\section{Pipeline schedule details}
\label{app:pipeline-schedules}
\label{app:noncolocated-schedule}

Section~\ref{sec:method-schedules} describes pipeline orchestration at the abstraction level.
This appendix gives the execution details for both placement modes.
In both cases, the boundary communicator defines the layout transform for an edge, while the schedule determines when that edge is active and which ranks must participate.

\begin{figure}[t]
  \centering
  \includegraphics[width=\linewidth]{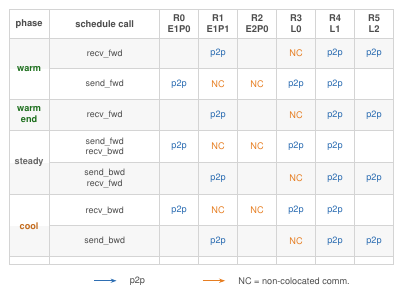}
  \caption{Graph-aware communication dispatch for the non-colocated topology in Figure~\ref{fig:pipeline-orchestration}(a). Rows are pipeline schedule calls during warmup, steady state, and cooldown; columns are rank roles in the expanded module graph. Each nonempty cell shows the communicator selected for the active edge: \texttt{p2p} for module-internal pipeline communication and \texttt{NC} for an encoder--LLM boundary edge using the non-colocated communicator. Paired steady-state calls resolve the forward and backward halves independently, so different halves of the same schedule call may use different edge types.}
  \label{fig:app-noncolocated-dag-schedule}
\end{figure}

For non-colocated execution, the scheduler extends ordinary 1F1B by resolving each pipeline action to an edge in the expanded module graph.
Figure~\ref{fig:app-noncolocated-dag-schedule} expands the topology in Figure~\ref{fig:pipeline-orchestration}(a), where Encoder~1 has two PP stages, Encoder~2 has one PP stage, and the LLM has three PP stages.
Rows correspond to the communication calls issued during warmup, steady state, and cooldown; columns correspond to rank roles in the expanded graph.
A \texttt{p2p} cell means that the active edge is internal to one module and uses that module's ordinary pipeline communicator.
An \texttt{NC} cell means that the active edge crosses from an encoder rank set to the LLM rank set and therefore invokes the edge-local non-colocated communicator.
Blank cells indicate that the rank has no active neighbor for that call.

Warmup illustrates why the dispatch must be graph-aware.
Encoder~1 first sends activations from E1P0 to E1P1 through module-local P2P; its next forward send crosses the encoder--LLM boundary and switches to the non-colocated communicator.
Encoder~2 has only one PP stage, so its first forward send is already a boundary transfer.
LLM PP0 receives the two encoder boundary tensors through their corresponding non-colocated communicators and becomes runnable for a microbatch only after the required activations have been materialized in the LLM layout.
The remaining LLM stages then proceed with ordinary LLM pipeline P2P.

The steady-state rows apply the same dispatch rule to paired 1F1B calls.
A call such as \texttt{send\_fwd}/\texttt{recv\_bwd} contains a forward action for a newer microbatch and a backward action for an older microbatch.
The runtime resolves the active edge for each half independently.
Thus, different ranks in the same schedule phase may use different communication paths, and a boundary rank can send a new activation through an \texttt{NC} edge while receiving an older gradient through the reverse of an \texttt{NC} edge.
The scheduler selects the active edge and microbatch; the non-colocated communicator then performs the leader exchange, DP fan-in or fan-out, and local TP/CP broadcast described in Section~\ref{sec:method-non-colocated}.

Cooldown uses the same edge identities in reverse.
LLM gradients move by P2P until they reach LLM PP0.
Boundary gradients then return through the non-colocated communicator attached to the original encoder--LLM edge, while encoder-internal gradients continue through module-local P2P, such as from E1P1 back to E1P0.
This is the key invariant at graph joins: even though LLM PP0 consumes multiple encoder activations for the same microbatch, backward routes each gradient through the boundary communicator associated with the forward edge that produced that activation.

For colocated execution, the schedule detail is different because there is no inter-rank boundary P2P between modules.
The issue is ordering.
Encoder execution may require collectives over the encoder layout, while the LLM 1F1B schedule staggers ranks across pipeline stages and microbatches.
The implementation therefore uses the three-phase schedule shown in Figure~\ref{fig:pipeline-orchestration}(b).
Phase 1 runs encoder forward once over the local microbatch group for the current LLM pipeline window, then applies the colocated forward transform to produce a packed boundary tensor in the LLM PP0 layout.
The runtime records the metadata needed for the reverse colocated transform, including the source and destination layout coordinates and the batch intervals used by any fan-in or fan-out transform.

Phase 2 runs the ordinary LLM pipeline schedule on detached boundary views.
The packed boundary tensor is detached as a leaf variable for the LLM phase and exposed to LLM PP0 as per-microbatch views.
As 1F1B advances, PP0 selects the view for the active microbatch, inserts it into that microbatch's text embeddings, and then uses standard LLM pipeline P2P for the remaining LLM stages.
During this phase, the LLM does not enter encoder collectives, which avoids mixing encoder-layout collectives with the staggered LLM pipeline order.

Phase 3 reconnects the LLM phase to the encoder side.
After the LLM pipeline drains, gradients accumulated on the packed boundary tensor are handed to the colocated backward transform.
If the boundary used fan-in, the transform selects from the destination-gradient shard the batch interval contributed by each source rank.
If the boundary used fan-out, sibling destination-gradient slices are all-gathered and concatenated to reconstruct the original source-layout gradient shard.
The runtime then invokes encoder backward over the saved batched encoder activations for the local microbatch group.

The detach in Phase 2 is therefore only a scheduling device, not a change in training semantics.
The LLM treats the packed boundary tensor as the leaf input to its pipeline phase, while the encoder-side graph, saved activations, and communicator metadata remain live.
The explicit gradient handoff at the start of Phase 3 supplies the boundary gradients needed by the colocated backward transform and reconnects them to the encoder backward computation.

The phase boundary also enables encoder-state offload.
Encoder parameters are needed during Phase 1 and Phase 3, but not while Phase 2 executes the detached LLM pipeline.
When offload is enabled, the runtime starts the device-to-host transfer after encoder forward, overlaps it with the LLM warmup and steady-state pipeline, starts the host-to-device reload during LLM cooldown, and synchronizes before encoder backward consumes the parameters.
Optimizer state can use the same split lifecycle when enabled, with transfers scheduled around the encoder forward/backward phases rather than inside the LLM pipeline loop.

\FloatBarrier

%% file: tex/appendix_throughput_tables.tex
\clearpage
\begingroup
\captionsetup{hypcap=false}

\section{Appendix}
\label{app:throughput-results}

This appendix reports the layouts, peak memory, step times, tokens/s gains, and TFLOPS/GPU gains
supporting Section~\ref{sec:experiments}.  All parallelism columns use
\(\mathrm{TP}/\mathrm{CP}/\mathrm{PP}/\mathrm{DP}\).  Encoder rows use
\(\mathrm{CP}=1\) and \(\mathrm{PP}=1\) unless otherwise stated.  Metric columns
report percentage gains over homogeneous, and step-time columns report absolute
milliseconds.  Memory columns report peak max-rank memory in GB.  Unless stated otherwise, LLM recompute is off.  For the
encoder-size sweeps, 1B encoder rows run without encoder recompute; larger
encoder rows use full encoder recompute.

\subsection*{Notation and acronyms}

\begin{center}
  \captionof{table}{Notation and acronyms used in the throughput tables.}
  \label{tab:app-notation}
  \small
  \setlength{\tabcolsep}{6pt}
  \renewcommand{\arraystretch}{1.08}
  \begin{tabular}{l l}
    \toprule
    \textbf{Symbol} & \textbf{Meaning} \\
    \midrule
    H & Homogeneous colocated execution \\
    \textbf{C} & Heterogeneous colocated execution \\
    NC & Heterogeneous non-colocated execution \\
    H-E / H-L & Encoder / LLM layout in an H run \\
    H Layout & Shared encoder / LLM layout in an H run \\
    \textbf{C}-E / \textbf{C}-L & Encoder / LLM layout in a \textbf{C} run \\
    NC-E / NC-L & Encoder / LLM layout in an NC run \\
    TP / CP / PP / DP & Tensor / context / pipeline / data parallelism \\
    TP/CP/PP/DP & Layout tuple order used in all tables \\
    SP & Sequence parallelism \\
    Seq. Len. & LLM sequence length \\
    Vis\% & Vision-token percentage \\
    GBS & Global batch size \\
    NMB & Number of microbatches \\
    Tokens/s & Aggregate training throughput in tokens/s \\
    TFLOPS/GPU & Per-GPU tera floating-point operations per second \\
    Peak Mem (GB) & Peak max-rank GPU memory in GB \\
    Step (ms) & Iteration step time in milliseconds \\
    F+B (ms) & Forward plus backward time in milliseconds \\
    Enc. Rec. / LLM Rec. & Encoder / LLM recomputation setting \\
    Enc. opt. inst. & Number of encoder distributed optimizer instances \\
    W5/S15 & Five warmup steps followed by fifteen measured steps \\
    \bottomrule
  \end{tabular}
\end{center}

\vspace{0.4em}

\noindent\textbf{Compute infrastructure.}
The empirical results reported in this paper were measured on a GPU cluster with
NVIDIA H100 80GB GPUs.  Cluster nodes contain eight GPUs; the throughput tables
report world size directly in GPUs.
Within a node, GPUs are connected with fourth-generation NVLink, providing
900~GB/s bidirectional GPU-to-GPU NVLink bandwidth with 18 NVLink connections
per GPU.  Across nodes, each node uses eight 400~Gb/s ConnectX-7 InfiniBand
cards.  This hardware setup is used for all empirical measurements reported in
the main text and for the tokens/s gain, TFLOPS/GPU gain, and step-time
measurements reported below.

\clearpage

\subsection{Colocated results}
\label{app:colocated-throughput-results}

\begin{samepage}
\begin{center}
  \captionof{table}{Colocated 8K gain, memory, and step-time results. GBS is
  288 for the 72-GPU rows and 256 for the tuned 256-GPU 70B/120B rows.}
  \label{tab:app-colocated-raw}
  \scriptsize
  \setlength{\tabcolsep}{2pt}
  \renewcommand{\arraystretch}{1.10}
  \makebox[\textwidth][c]{\resizebox{\textwidth}{!}{%
  \begin{tabular}{l c c l l l l c c c c}
    \toprule
    \multirow{2}{*}{\textbf{Model}} &
    \multirow{2}{*}{\textbf{Vis\%}} &
    \multirow{2}{*}{\textbf{GPUs}} &
    \textbf{H-E} & \textbf{H-L} & \textbf{C-E} & \textbf{C-L} &
    \textbf{Enc. opt. inst.}\textsuperscript{*} & \textbf{TFLOPS/GPU Gain} & \textbf{Peak Mem (GB)} & \textbf{Step (ms)} \\
    & & & TP/CP/PP/DP & TP/CP/PP/DP & TP/CP/PP/DP & TP/CP/PP/DP &
    & \(\Delta\%\) & H / \textbf{C} & H / \textbf{C} \\
    \midrule
    1B+14B & 50   & 72 & 4/1/1/18 & 4/1/1/18 & 2/1/1/36 & 2/1/2/18 & 1 & \(\mathbf{+12.1}\) & 57.2 / 57.1 & 7,570 / \textbf{6,750} \\
    1B+14B & 75   & 72 & 4/1/1/18 & 4/1/1/18 & 1/1/1/72 & 2/1/2/18 & 1 & \(\mathbf{+13.5}\) & 57.2 / 57.6 & 7,780 / \textbf{6,840} \\
    1B+14B & 87.5 & 72 & 4/1/1/18 & 4/1/1/18 & 1/1/1/72 & 2/1/2/18 & 1 & \(\mathbf{+12.7}\) & 57.2 / 57.8 & 7,660 / \textbf{6,800} \\
    3B+14B & 50   & 72 & 4/1/1/18 & 4/1/1/18 & 2/1/1/36 & 2/1/2/18 & 1 & \(\mathbf{+12.9}\) & 59.7 / 57.3 & 7,850 / \textbf{6,970} \\
    3B+14B & 75   & 72 & 4/1/1/18 & 4/1/1/18 & 1/1/1/72 & 2/1/2/18 & 1 & \(\mathbf{+12.7}\) & 59.7 / 57.7 & 8,020 / \textbf{7,140} \\
    3B+14B & 87.5 & 72 & 4/1/1/18 & 4/1/1/18 & 1/1/1/72 & 2/1/2/18 & 1 & \(\mathbf{+14.0}\) & 59.7 / 57.9 & 8,130 / \textbf{7,150} \\
    \addlinespace[2pt]
    1B+32B & 75   & 72 & 4/1/1/18 & 4/1/1/18 & 1/1/1/72 & 4/1/2/9  & 1 & on par & 66.2 / 55.2 & \textbf{14,210} / 14,380 \\
    3B+32B & 50   & 72 & 4/1/1/18 & 4/1/1/18 & 1/1/1/72 & 4/1/2/9  & 1 & on par & 68.7 / 54.2 & \textbf{14,330} / 14,750 \\
    3B+32B & 75   & 72 & 4/1/1/18 & 4/1/1/18 & 1/1/1/72 & 4/1/2/9  & 1 & \(\mathbf{+11.1}\) & 68.7 / 55.3 & 14,370 / \textbf{12,950} \\
    3B+32B & 87.5 & 72 & 4/1/1/18 & 4/1/1/18 & 1/1/1/72 & 4/1/2/9  & 1 & \(\mathbf{+13.1}\) & 68.7 / 55.8 & 14,630 / \textbf{12,910} \\
    \addlinespace[2pt]
    1B+70B & 50 & 256 & 8/1/1/32 & 8/1/1/32 & 1/1/1/256 & 8/1/2/16 & 4 & \(\mathbf{+16.5}\) & 61.5 / 47.2 & 8,441 / \textbf{7,244} \\
    1B+70B & 75 & 256 & 8/1/1/32 & 8/1/1/32 & 1/1/1/256 & 8/1/2/16 & 4 & \(\mathbf{+15.1}\) & 61.5 / 47.8 & 8,415 / \textbf{7,312} \\
    3B+70B & 50 & 256 & 8/1/1/32 & 8/1/1/32 & 1/1/1/256 & 8/1/2/16 & 4 & \(\mathbf{+13.9}\) & 62.7 / 48.8 & 8,471 / \textbf{7,434} \\
    3B+70B & 75 & 256 & 8/1/1/32 & 8/1/1/32 & 1/1/1/256 & 8/1/2/16 & 4 & \(\mathbf{+12.7}\) & 62.7 / 49.4 & 8,452 / \textbf{7,501} \\
    1B+120B & 50 & 256 & 8/1/1/8 & 8/1/4/8 & 1/1/1/256 & 8/1/4/8 & 8 & \(\mathbf{+16.6}\) & 57.2 / 54.0 & 15,258 / \textbf{13,081} \\
    1B+120B & 75 & 256 & 8/1/1/8 & 8/1/4/8 & 1/1/1/256 & 8/1/4/8 & 8 & \(\mathbf{+13.9}\) & 63.2 / 55.4 & 15,169 / \textbf{13,319} \\
    3B+120B & 50 & 256 & 8/1/1/8 & 8/1/4/8 & 2/1/1/128 & 8/1/4/8 & 2 & \(\mathbf{+21.5}\) & 60.5 / 66.2 & 15,093 / \textbf{12,420} \\
    3B+120B & 75 & 256 & 8/1/1/8 & 8/1/4/8 & 2/1/1/128 & 8/1/4/8 & 2 & \(\mathbf{+21.6}\) & 60.5 / 67.6 & 15,303 / \textbf{12,579} \\
    6B+120B & 50 & 256 & 8/1/1/8 & 8/1/4/8 & 2/1/1/128 & 8/1/4/8 & 1 & \(-0.7\) & 53.6 / 68.0 & \textbf{17,890} / 18,020 \\
    7.2B+120B & 50 & 256 & 8/1/1/8 & 8/1/4/8 & 4/1/1/64 & 8/1/4/8 & 1 & \(-0.3\) & 53.6 / 64.4 & \textbf{17,980} / 18,040 \\
    12B+120B & 50 & 256 & 8/1/1/8 & 8/1/4/8 & 4/1/1/64 & 8/1/4/8 & 1 & \(-8.0\) & 66.9 / 69.6 & \textbf{15,860} / 18,190 \\
    \bottomrule
  \end{tabular}
  }}
\end{center}
\end{samepage}

\begingroup
\renewcommand{\thefootnote}{\fnsymbol{footnote}}
\footnotetext[1]{Enc. opt. inst. reports the number of encoder distributed
optimizer instances when swept; default is 1.}
\endgroup

\vspace{0.2em}
\noindent
The 8K colocated sweep shows where module-local layout freedom is useful on the
same GPUs. The 14B LLM rows gain 12.1\%--14.0\% TFLOPS/GPU; for 32B, gains
concentrate at higher vision-token share. The tuned
256-GPU rows show a second colocated operating point: reducing encoder TP,
adding LLM pipeline parallelism where useful, and tuning encoder optimizer
instances gives +12.7\%--21.6\% for the audited 1B/3B encoder rows with 70B and
120B LLMs. For larger 120B encoder rows, \textbf{C} converges back toward H or
slows down, motivating the NC crossover summarized in
Table~\ref{tab:app-operating-ranges}.

\begin{center}
  \captionof{table}{Colocated long-context gain, memory, and step-time results.
  Encoder CP remains 1 under heterogeneous parallelism, while homogeneous
  training forces the encoder to inherit LLM CP.  F+B is forward plus backward
  time in ms. Because H and \textbf{C} use the same GPU count, tokens/s and
  TFLOPS/GPU gains are identical; Gain reports both.}
  \label{tab:app-long-context-raw}
  \scriptsize
  \setlength{\tabcolsep}{1.5pt}
  \renewcommand{\arraystretch}{1.00}
  \makebox[\textwidth][c]{\resizebox{\textwidth}{!}{%
  \begin{tabular}{l c c c c c l l l c c c}
    \toprule
    \multirow{2}{*}{\textbf{Model}} &
    \multirow{2}{*}{\textbf{Seq. Len.}} &
    \multirow{2}{*}{\textbf{Vis\%}} &
    \multirow{2}{*}{\textbf{GPUs}} &
    \multirow{2}{*}{\textbf{GBS}} &
    \multirow{2}{*}{\textbf{NMB}} &
    \textbf{H Layout} & \textbf{C-E} & \textbf{C-L} &
    \textbf{Gain} & \textbf{Peak Mem (GB)} & \textbf{F+B (ms)} \\
    & & & & & & TP/CP/PP/DP & TP/CP/PP/DP & TP/CP/PP/DP &
    \(\Delta\%\) & H / \textbf{C} & H / \textbf{C} \\
    \midrule
    1B+7B  & 16K  & 50 & 16 & 8 & 2 & 2/2/1/4  & 1/1/1/16 & 2/2/1/4  & \(\mathbf{+6.6}\)  & 54.5 / 50.2 & 1,300 / \textbf{1,221} \\
    1B+7B  & 16K  & 75 & 16 & 8 & 2 & 2/2/1/4  & 1/1/1/16 & 2/2/1/4  & \(\mathbf{+8.5}\)  & 59.6 / 52.2 & 1,352 / \textbf{1,248} \\
    1B+7B  & 32K  & 50 & 16 & 4 & 2 & 2/4/1/2  & 1/1/1/16 & 2/4/1/2  & \(\mathbf{+11.7}\) & 65.3 / 50.7 & 1,836 / \textbf{1,644} \\
    1B+7B  & 32K  & 75 & 16 & 4 & 2 & 2/4/1/2  & 1/1/1/16 & 2/4/1/2  & \(\mathbf{+20.3}\) & 75.6 / 52.7 & 1,980 / \textbf{1,640} \\
    1B+7B  & 64K  & 50 & 32 & 4 & 2 & 4/8/1/1  & 1/1/1/32 & 2/8/1/2  & \(\mathbf{+34.3}\) & 54.0 / 48.4 & 3,966 / \textbf{2,937} \\
    1B+7B  & 64K  & 75 & 32 & 4 & 2 & 4/8/1/1  & 1/1/1/32 & 2/8/1/2  & \(\mathbf{+49.3}\) & 68.9 / 50.3 & 4,457 / \textbf{2,973} \\
    1B+7B  & 128K & 50 & 64 & 4 & 2 & 4/16/1/1 & 1/1/1/64 & 2/16/1/2 & \(\mathbf{+17.8}\) & 25.1 / 48.6 & 6,321 / \textbf{5,381} \\
    1B+7B  & 128K & 75 & 64 & 4 & 2 & 4/16/1/1 & 1/1/1/64 & 2/16/1/2 & \(\mathbf{+24.8}\) & 26.6 / 50.4 & 6,614 / \textbf{5,291} \\
    \addlinespace[2pt]
    1B+14B & 16K  & 50 & 16 & 4 & 1 & 4/1/1/4  & 2/1/1/8  & 4/1/1/4  & \(\mathbf{+0.1}\)  & 75.1 / 73.8 & \textbf{1,202} / 1,205 \\
    1B+14B & 16K  & 75 & 16 & 4 & 1 & 8/1/1/2  & 2/1/1/8  & 4/1/1/4  & \(\mathbf{+16.3}\) & 50.4 / 76.3 & 1,438 / \textbf{1,211} \\
    1B+14B & 32K  & 50 & 32 & 4 & 1 & 4/2/1/4  & 2/1/1/16 & 4/2/1/4  & \(\mathbf{+2.9}\)  & 76.6 / 67.8 & 1,628 / \textbf{1,583} \\
    1B+14B & 32K  & 75 & 32 & 4 & 1 & 8/2/1/2  & 2/1/1/16 & 4/2/1/4  & \(\mathbf{+21.3}\) & 53.5 / 70.3 & 1,962 / \textbf{1,585} \\
    1B+14B & 64K  & 50 & 32 & 4 & 2 & 8/4/1/1  & 2/1/1/16 & 4/4/1/2  & \(\mathbf{+23.9}\) & 61.2 / 69.1 & 5,687 / \textbf{4,579} \\
    1B+14B & 64K  & 75 & 32 & 4 & 2 & 8/4/1/1  & 2/1/1/16 & 4/4/1/2  & \(\mathbf{+30.3}\) & 73.2 / 71.6 & 5,848 / \textbf{4,479} \\
    1B+14B & 128K & 50 & 64 & 4 & 2 & 4/8/2/1  & 2/1/1/32 & 4/8/1/2  & \(\mathbf{+29.8}\) & 55.4 / 63.0 & 9,397 / \textbf{7,184} \\
    1B+14B & 128K & 75 & 64 & 4 & 2 & 4/8/2/1  & 2/1/1/32 & 4/8/1/2  & \(\mathbf{+34.0}\) & 55.8 / 62.9 & 9,692 / \textbf{7,175} \\
    \bottomrule
  \end{tabular}
  }}
\end{center}

\vspace{0.2em}
\noindent
Long context gives the clearest case for \textbf{C}: the LLM needs context
parallelism as sequence length grows, but the encoder sequence length is fixed
by image resolution rather than LLM context. H makes the encoder inherit LLM CP,
while \textbf{C} keeps the encoder at CP=1 and converts those ranks into encoder
DP replicas. Gains therefore grow with context, reaching 34.3\%--49.3\% for
1B+7B at 64K and 29.8\%--34.0\% for 1B+14B at 128K.

\clearpage

\subsection{Non-colocated results}
\label{app:noncolocated-throughput-results}

\begin{center}
  \captionof{table}{Non-colocated 120B LLM gain, memory, and step-time results. Rows compare
  homogeneous shared-grid and non-colocated (NC) execution with a dedicated
  encoder island. H Layout is the shared encoder/LLM layout for the H run. Rows
  are grouped by encoder size. All rows use sequence parallelism (SP) on and LLM
  recompute off; encoder full recompute is enabled.}
  \label{tab:app-noncolocated-raw}
  \vspace{0.15em}
  \scriptsize
  \setlength{\tabcolsep}{1.4pt}
  \renewcommand{\arraystretch}{0.98}
  \makebox[\textwidth][c]{\resizebox{\textwidth}{!}{%
  \begin{tabular}{l c c c c l l l c c c c}
    \toprule
    \multirow{2}{*}{\textbf{Model}} &
    \multirow{2}{*}{\textbf{Seq. Len.}} &
    \multirow{2}{*}{\textbf{Vis\%}} &
    \multirow{2}{*}{\textbf{GPUs}} &
    \multirow{2}{*}{\textbf{GBS}} &
    \textbf{H Layout} & \textbf{NC-E} & \textbf{NC-L} &
    \textbf{Tokens/s Gain} & \textbf{TFLOPS/GPU Gain} & \textbf{Peak Mem (GB)} & \textbf{Step (ms)} \\
    & & & & & TP/CP/PP/DP & TP/CP/PP/DP & TP/CP/PP/DP &
    \(\Delta\%\) & \(\Delta\%\) & H / NC & H / NC \\
    \midrule
    6B+120B & 8K & 50 & 256/264 & 256 & 8/1/4/8 & 1/1/1/8 & 8/1/2/16 & \(\mathbf{+5.9}\) & \(\mathbf{+2.7}\) & 64.8 / 76.6 & 17,700 / \textbf{16,710} \\
    6B+120B & 8K & 75 & 256/264 & 256 & 8/1/4/8 & 1/1/1/8 & 8/1/2/16 & \(\mathbf{+4.4}\) & \(\mathbf{+1.3}\) & 53.6 / 76.8 & 17,600 / \textbf{16,860} \\
    6B+120B & 8K & 50 & 512/520 & 512 & 8/1/4/16 & 1/1/1/8 & 8/1/2/32 & \(\mathbf{+4.0}\) & \(\mathbf{+2.4}\) & 50.4 / 73.5 & 17,870 / \textbf{17,180} \\
    6B+120B & 16K & 50 & 256/264 & 256 & 8/2/4/4 & 1/1/1/8 & 8/2/4/4 & \(\mathbf{+0.7}\) & \(-2.3\) & 53.6 / 53.6 & 38,380 / \textbf{38,110} \\
    6B+120B & 16K & 75 & 256/264 & 256 & 8/2/4/4 & 2/1/1/4 & 8/2/4/4 & \(\mathbf{+3.3}\) & \(\mathbf{+0.2}\) & 53.6 / 63.8 & 38,120 / \textbf{36,900} \\
    6B+120B & 16K & 50 & 512/520 & 512 & 8/2/4/8 & 1/1/1/8 & 8/2/2/16 & \(\mathbf{+5.1}\) & \(\mathbf{+3.5}\) & 50.4 / 74.6 & 38,400 / \textbf{36,550} \\
    \addlinespace[2pt]
    7.2B+120B & 8K & 50 & 256/264 & 256 & 8/1/4/8 & 1/1/1/8 & 8/1/2/16 & \(\mathbf{+7.1}\) & \(\mathbf{+3.8}\) & 53.6 / 76.6 & 17,980 / \textbf{16,790} \\
    7.2B+120B & 8K & 75 & 256/264 & 256 & 8/1/2/16 & 1/1/1/8 & 8/1/2/16 & \(\mathbf{+13.0}\) & \(\mathbf{+9.6}\) & 77.2 / 73.3 & 16,210 / \textbf{14,340} \\
    7.2B+120B & 16K & 50 & 256/264 & 256 & 8/2/4/4 & 1/1/1/8 & 8/2/4/4 & \(\mathbf{+5.5}\) & \(\mathbf{+2.3}\) & 53.6 / 63.2 & 38,260 / \textbf{36,260} \\
    7.2B+120B & 16K & 75 & 256/264 & 256 & 8/2/4/4 & 2/1/1/4 & 8/2/4/4 & \(\mathbf{+3.6}\) & \(\mathbf{+0.4}\) & 53.6 / 63.8 & 38,380 / \textbf{37,070} \\
    7.2B+120B & 16K & 50 & 512/520 & 512 & 8/2/4/8 & 1/1/1/8 & 8/2/2/16 & \(\mathbf{+4.9}\) & \(\mathbf{+3.3}\) & 50.4 / 74.6 & 38,450 / \textbf{36,650} \\
    \addlinespace[2pt]
    12B+120B & 8K & 50 & 256/264 & 256 & 8/1/4/8 & 2/1/1/4 & 8/1/2/16 & \(\mathbf{+7.3}\) & \(\mathbf{+4.1}\) & 66.9 / 73.1 & 15,860 / \textbf{14,780} \\
    12B+120B & 8K & 75 & 256/264 & 256 & 8/1/4/8 & 2/1/1/4 & 8/1/2/16 & \(\mathbf{+12.7}\) & \(\mathbf{+9.3}\) & 67.0 / 73.3 & 16,690 / \textbf{14,810} \\
    12B+120B & 8K & 87.5 & 256/264 & 256 & 8/1/4/8 & 2/1/1/4 & 8/1/2/16 & \(\mathbf{+9.5}\) & \(\mathbf{+6.2}\) & 67.0 / 73.4 & 17,020 / \textbf{15,540} \\
    12B+120B & 16K & 50 & 256/264 & 256 & 8/2/4/4 & 2/1/1/4 & 8/2/4/4 & \(\mathbf{+5.6}\) & \(\mathbf{+2.4}\) & 56.3 / 63.2 & 38,430 / \textbf{36,400} \\
    12B+120B & 16K & 50 & 512/520 & 512 & 8/2/4/8 & 2/1/1/4 & 8/2/2/16 & \(\mathbf{+2.5}\) & \(\mathbf{+0.9}\) & 52.6 / 74.6 & 38,370 / \textbf{37,430} \\
    \bottomrule
  \end{tabular}
  }}
\end{center}

\vspace{0.2em}
\noindent
The 120B NC rows show physical separation becoming useful once encoder state
constrains the LLM layout. At 8K and 50\% vision tokens, NC is positive from 6B
onward and sharpens for the 7.2B and 12B encoders; at 16K, the 512/520-GPU rows show the
scale-out benefit of preserving the LLM
island. Because NC averages over the extra 8-GPU encoder island, TFLOPS/GPU gains
are conservative relative to the aggregate tokens/s gains.

\begin{center}
  \begin{minipage}[t]{0.49\textwidth}
    \centering
    \includegraphics[width=\linewidth]{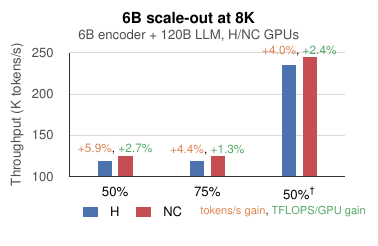}
    {\footnotesize\textbf{(a)} 6B+120B 8K scale-out}
  \end{minipage}
  \hfill
  \begin{minipage}[t]{0.49\textwidth}
    \centering
    \includegraphics[width=\linewidth]{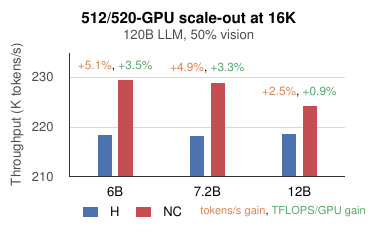}
    {\footnotesize\textbf{(b)} 120B LLM 16K scale-out}
  \end{minipage}
  \captionof{figure}{Non-colocated 120B scale-out results. (a) Non-colocation remains positive for the 6B+120B case when expanding from 256/264 to 512/520 GPUs. (b) At 512/520 GPUs, NC preserves the PP2 LLM layout and gives +5.1\%, +4.9\%, and +2.5\% tokens/s gains for 6B, 7.2B, and 12B encoders. Labels report tokens/s and TFLOPS/GPU gains over H.}
  \label{fig:app-noncolocated-scaleout}
\end{center}

\subsection{Operating ranges for different modes}
\label{app:operating-ranges}

\begin{center}
  \captionof{table}{H/\textbf{C}/NC operating-range comparison at 8K. \textbf{C}
  and NC gains are relative to H for the same workload; \(\textbf{C}\)
  vs NC is the gap of \textbf{C} relative to NC. Each gain cell reports
  tokens/s / TFLOPS/GPU. When \textbf{C} converges to
  H, we report \textbf{C}=H because H is a valid special case of colocated
  heterogeneous execution. Small-encoder rows use 72
  total GPUs for all modes, with NC splitting 64 LLM GPUs plus an 8-GPU encoder
  island. The 70B/120B rows use 256 GPUs for H/\textbf{C} and 264 GPUs for NC.
  The 70B/120B \textbf{C} rows report the best tuned colocated configuration,
  including encoder offload where available.}
  \label{tab:app-operating-ranges}
  \scriptsize
  \setlength{\tabcolsep}{2pt}
  \renewcommand{\arraystretch}{1.06}
  \makebox[\textwidth][c]{\resizebox{\textwidth}{!}{%
  \begin{tabular}{l c c c c c c c c c}
    \toprule
    \textbf{Model} & \textbf{Vis\%} & \textbf{GPUs} &
    \textbf{C-E} & \textbf{NC-E} &
    \textbf{Enc. opt. inst.}\textsuperscript{*} & \textbf{C Gain} & \textbf{NC Gain} &
    \textbf{C vs NC} &
    \textbf{Best} \\
    & & H / \textbf{C} / NC &
    TP/CP/PP/DP & TP/CP/PP/DP &
    & tokens/s / TFLOPS/GPU & tokens/s / TFLOPS/GPU & tokens/s / TFLOPS/GPU & \\
    \midrule
    1B+14B   & 50   & 72 / 72 / 72    & 2/1/1/36  & 1/1/1/8 & 1 & \textbf{+12.1 / +12.1} & -11.3 / -11.3 & +26.4 / +26.3 & \textbf{C} \\
    1B+14B   & 87.5 & 72 / 72 / 72    & 1/1/1/72  & 1/1/1/8 & 1 & \textbf{+12.7 / +12.7} & -10.5 / -10.5 & +25.9 / +25.8 & \textbf{C} \\
    3B+14B   & 50   & 72 / 72 / 72    & 2/1/1/36  & 1/1/1/8 & 1 & \textbf{+12.9 / +12.9} & -8.6 / -8.6  & +23.5 / +23.5 & \textbf{C} \\
    3B+14B   & 87.5 & 72 / 72 / 72    & 1/1/1/72  & 1/1/1/8 & 1 & \textbf{+14.0 / +14.0} & -5.4 / -5.4  & +20.5 / +20.6 & \textbf{C} \\
    1B+32B   & 75   & 72 / 72 / 72    & 1/1/1/72  & 1/1/1/8 & 1 & on par / -0.9 & -10.2 / -10.2 & +11.4 / +10.4 & H/\textbf{C} \\
    3B+32B   & 50   & 72 / 72 / 72    & 1/1/1/72  & 1/1/1/8 & 1 & on par / -0.1 & -10.1 / -10.1 & +11.2 / +11.1 & H/\textbf{C} \\
    3B+32B   & 87.5 & 72 / 72 / 72    & 1/1/1/72  & 1/1/1/8 & 1 & \textbf{+13.1 / +13.1} & -7.8 / -7.8  & +22.7 / +22.7 & \textbf{C} \\
    \addlinespace[2pt]
    1B+70B   & 50   & 256 / 256 / 264 & 1/1/1/256 & 1/1/1/8 & 4  & \textbf{+16.5 / +16.5} & +7.3 / +4.0 & +8.6 / +12.0 & \textbf{C} \\
    3B+70B   & 50   & 256 / 256 / 264 & 1/1/1/256 & 1/1/1/8 & 4  & \textbf{+13.9 / +13.9} & +9.3 / +6.0 & +4.2 / +7.5 & \textbf{C} \\
    \addlinespace[2pt]
    1B+120B  & 50   & 256 / 256 / 264 & 1/1/1/256 & 1/1/1/8 & 8  & \textbf{+16.6 / +16.6} & -3.0 / -0.5 & +20.2 / +17.2 & \textbf{C} \\
    3B+120B  & 50   & 256 / 256 / 264 & 2/1/1/128 & 1/1/1/8 & 2  & \textbf{+21.5 / +21.5} & -9.9 / -7.7 & +34.9 / +31.6 & \textbf{C} \\
    6B+120B  & 50   & 256 / 256 / 264 & H layout  & 1/1/1/8 & 1 & \textbf{C}=H  & \textbf{+5.9 / +2.7} & -5.6 / -2.6  & NC \\
    7.2B+120B & 50  & 256 / 256 / 264 & H layout  & 1/1/1/8 & 1 & \textbf{C}=H  & \textbf{+7.1 / +3.8} & -6.6 / -3.7  & NC \\
    12B+120B & 50   & 256 / 256 / 264 & H layout  & 2/1/1/4 & 1 & \textbf{C}=H  & \textbf{+7.3 / +4.1} & -6.8 / -3.9  & NC \\
    \bottomrule
  \end{tabular}
  }}
\end{center}

\begingroup
\renewcommand{\thefootnote}{\fnsymbol{footnote}}
\footnotetext[1]{Enc. opt. inst. reports the number of encoder distributed
optimizer instances when swept; default is 1.}
\endgroup

\vspace{0.15em}
\noindent
Table~\ref{tab:app-operating-ranges} shows the operating ranges of H,
\textbf{C}, and NC. When the combined model has enough memory headroom,
\textbf{C} is the best use of the fixed GPU budget: it remaps the encoder to a
module-local layout with more data parallelism while keeping the LLM colocated.
That larger encoder-DP group increases cross-node optimizer communication, so
the tuned \textbf{C} rows use multiple encoder optimizer instances where memory
allows. Because encoder parameters can be offloaded, the added optimizer-state
replication is affordable and improves throughput. As LLM and encoder memory
pressure increase, this colocated flexibility diminishes: \textbf{C} converges
back to H because H is a valid special case of \textbf{C}, and both modules
still share the same physical ranks. NC becomes preferable once the encoder and LLM should stop
carrying each other's memory and scheduling pressure. At the 120B LLM scale,
the verified 1B/3B rows still favor \textbf{C}, while the separation point appears
at the 6B encoder scale, or equivalent encoder-side compute and memory from
multiple encoders: from 6B+120B onward, the dedicated
encoder island unlocks the PP2 LLM layout and becomes the preferred operating
point.

\subsection{Bridge communication}
\label{app:bridge-communication}

\begin{center}
  \includegraphics[width=0.98\linewidth]{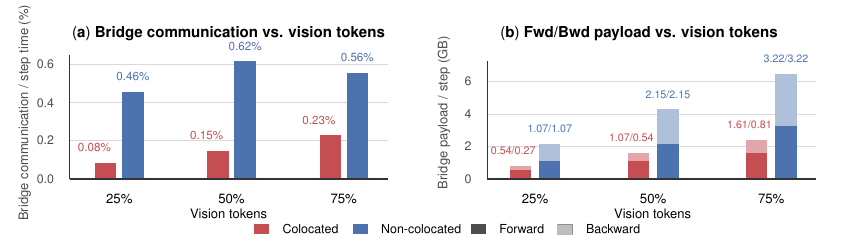}
  \captionof{figure}{Bridge communication profile for the 3B encoder + 70B LLM
  payload sweep at 8K sequence length. Panel (a) reports median bridge
  communication time as a percentage of measured step time. Panel (b) reports
  local bridge payload per step as F/B, where F is forward payload and B is
  backward payload, both in GB.}
  \label{fig:app-bridge-communication-profile}
\end{center}

\vspace{0.2em}
\noindent
In the 3B+70B colocated setup, the encoder uses \(\mathrm{DP}=32\) and the LLM
uses \(\mathrm{DP}=16\), so the fan-in scale is 2. Each colocated bridge group
is therefore a 2-rank all-gather, not a global DP32 or DP16 collective. This
small-group fan-in explains why colocated forward bridge time is low. NC
forward is larger because it includes both cross-island activation transfer and
LLM-side activation fanout: the encoder island sends activations to the LLM
receiver ranks, and those activations are then broadcast within the relevant
LLM-side group. NC backward is simpler in this setup because the encoder island
uses \(\mathrm{TP}=1\); there is no encoder-side TP broadcast, so the backward
bridge is the LLM-to-encoder gradient transport.

\endgroup
\clearpage